%% file: root.tex
\documentclass[letterpaper, 10 pt, journal, twoside]{IEEEtran}  

\usepackage[utf8]{inputenc}
\usepackage[T1]{fontenc}

\usepackage{microtype}

\usepackage[pdftex]{graphicx}
\usepackage{epstopdf}

\usepackage{amsmath} 
\usepackage{amssymb} 
\usepackage{etoolbox,siunitx}
\DeclareSIUnit[number-unit-product = ]\k{k} 
\robustify\bfseries
\robustify\itshape

\input{inc/mathstuff}

\usepackage{booktabs}
\usepackage{multirow}
\usepackage{multicol}
\usepackage[table,xcdraw]{xcolor}

\usepackage{paralist}

\usepackage[vlined,ruled]{algorithm2e}

\usepackage[caption=false, font=footnotesize]{subfig}

\usepackage[shortcuts]{extdash}

\usepackage{hyperref}
\hypersetup{
    pdftitle={A Credible and Robust approach to Ego-Motion Estimation using an Automotive Radar},
    pdfauthor={Karim Haggag, Sven Lange, Tim Pfeifer and Peter Protzel},
    pdfsubject={CREME -- Credible Radar Ego-Motion Estimation},
    pdfkeywords={Range Sensing, Probabilistic Inference, Probability and Statistical Methods, Radar Odometry, Scan Matching, Registration, Automotive Radar, Optimization, Gaussian Mixtures, Factor Graphs, Least Squares, Sensor Fusion, State Estimation}
}

\usepackage{pgfplots}
\usepgfplotslibrary{statistics}

\def\equationautorefname~#1\null{(#1)\null}

\usepackage{paralist} 

\usepackage{fancyhdr}
\begin{document}
    
\markboth{IEEE Robotics and Automation Letters. Preprint Version. Accepted March, 2022}
{}  

\author{Karim Haggag, Sven Lange, Tim Pfeifer and Peter Protzel
\thanks{Manuscript received: November 26, 2021; Revised February 11, 2022; Accepted March 14, 2022.}
\thanks{This paper was recommended for publication by Editor Sven Behnke upon evaluation of the Associate Editor and Reviewers' comments.}
\thanks{The authors are with the Faculty of Electrical Engineering and Information Technology,
            Technische Universität Chemnitz, Germany,\\
            (e-mail: {\tt\footnotesize \{firstname.lastname\}@etit.tu-chemnitz.de})}%
\thanks{Digital Object Identifier (DOI): \href{https://ieeexplore.ieee.org/document/9743799}{10.1109/LRA.2022.3162644}}
}

\title{A Credible and Robust approach to Ego-Motion Estimation using an Automotive Radar}

\maketitle

\fancyfoot{}
\fancyhead[OL]{ 
    \footnotesize
    Published in IEEE Robotics and Automation Letters (RA-L), 2022. ACCEPTED VERSION\\
    \tiny
    \copyright 2022 IEEE. Personal use of this material is permitted.  Permission from IEEE must be obtained for all other uses, in any current or future media, including reprinting/republishing this material for advertising or promotional purposes, creating new collective works, for resale or redistribution to servers or lists, or reuse of any copyrighted component of this work in other works.
}
\addtolength{\headheight}{\baselineskip}
\thispagestyle{fancy}
\pagestyle{empty}


\begin{abstract}

Consistent motion estimation is fundamental for all mobile autonomous systems.
While this sounds like an easy task, often, it is not the case because of changing environmental conditions affecting odometry obtained from vision, Lidar, or the wheels themselves.
Unsusceptible to challenging lighting and weather conditions, radar sensors are an obvious alternative.

Usually, automotive radars return a sparse point cloud, representing the surroundings.
Utilizing this information to motion estimation is challenging due to unstable and phantom measurements, which result in a high rate of outliers.

We introduce a credible and robust probabilistic approach to estimate the ego-motion based on these challenging radar measurements; intended to be used within a loosely-coupled sensor fusion framework.
Compared to existing solutions, evaluated on the popular nuScenes dataset and others, we show that our proposed algorithm is more credible while not depending on explicit correspondence calculation.

\end{abstract}


\input{inc/intro.tex}

\input{inc/prior.tex}

\input{inc/algorithm.tex}

\input{inc/eval.tex}

\input{inc/results.tex}

\input{inc/conclusion.tex}

\bibliographystyle{IEEEtran}
\bibliography{IEEEabrv,lasve}

\end{document}

%% file: inc/mathstuff.tex
\newcommand{\vect}[1]{\mathbf{ #1}}

\newcommand{\vectg}[1]{{\boldsymbol{ #1}}}

\newcommand{\T}{^\mathsf{T}}

\newcommand{\correspondto}{\mathrel{\widehat{=}}}

\newcommand{\diag}{\operatorname{diag}}

\newcommand{\argmin}{\operatornamewithlimits{argmin}}

\newcommand{\argmax}{\operatornamewithlimits{argmax}}

\newcommand{\normal}[2]{\mathcal{N}\left(#1\, |\, #2\right)}

\newcommand{\uniform}[2]{\mathcal{U}\left(#1\, |\, #2\right)}

\usepackage{xparse}
\NewDocumentCommand\PP{mg}{%
    \ensuremath{P\left( #1 \IfNoValueTF{#2}{}{\, |\, #2} \right)}%
}

\newcommand{\pfrac}[2]{\frac{\partial #1}{\partial #2}}  

\newcommand{\smd}[2]{\left\| #1 \right\|^2_{#2}}

\newcommand{\twovector}[2]{\begin{pmatrix} #1 \\ #2 \end{pmatrix}} 
 
\newcommand{\vL}{\vect{L}}

\newcommand{\vR}{\vect{R}}

\newcommand{\vm}{\vect{m}}

\newcommand{\vr}{\vect{r}}

\newcommand{\vx}{\vect{x}}

\newcommand{\vmu}{\vectg{\mu}}

\newcommand{\vrho}{\vectg{\rho}}

\newcommand{\vSigma}{\vectg{\Sigma}}

\newcommand{\cF}{\mathcal{F}}

\newcommand{\cM}{\mathcal{M}}

\DeclareMathAlphabet\mathbfcal{OMS}{cmsy}{b}{n}


%% file: inc/intro.tex

\section{Introduction}


\IEEEPARstart{I}{n}
order to develop autonomous vehicles, a robust and fault-aware localization or SLAM system is essential.
Consequently, the combination of various sensors with their specific strengths is mandatory.
Consider the following: a vehicle drives in an urban environment between tall buildings or inside tunnels on a rainy, foggy, or snowy day at night. 
A global navigation satellite system (GNSS) most likely will have non-line-of-sight and multipath errors.
Lidar suffers from rain and fog, and the vision sensors have poor accuracy because of the lighting conditions and the presence of snow.
However, a radar sensor would show a robust behavior within these challenging environmental conditions, and by using a Doppler radar, its detections are augmented by velocity information based on Doppler shift.

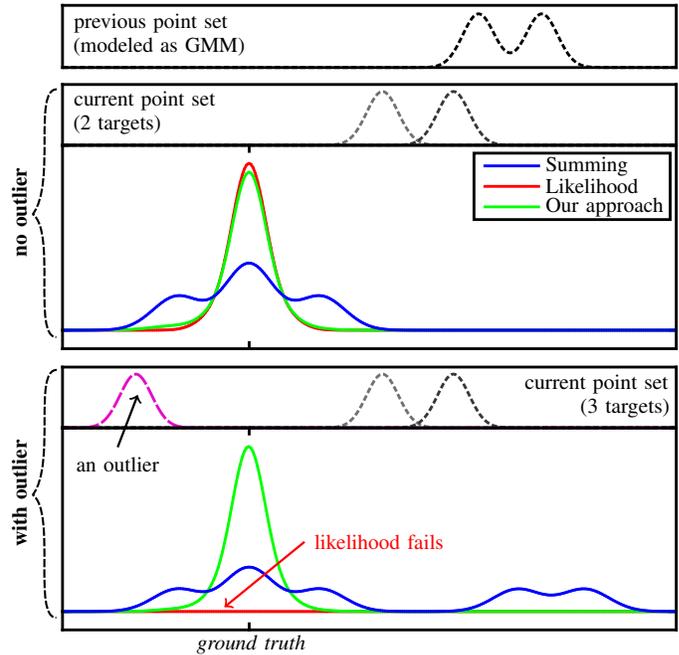
\begin{figure}[th]
\centering
\begin{tikzpicture}
  \node[inner sep=0pt, anchor=south west] (img) at (0,0){\includegraphics[width=\linewidth]{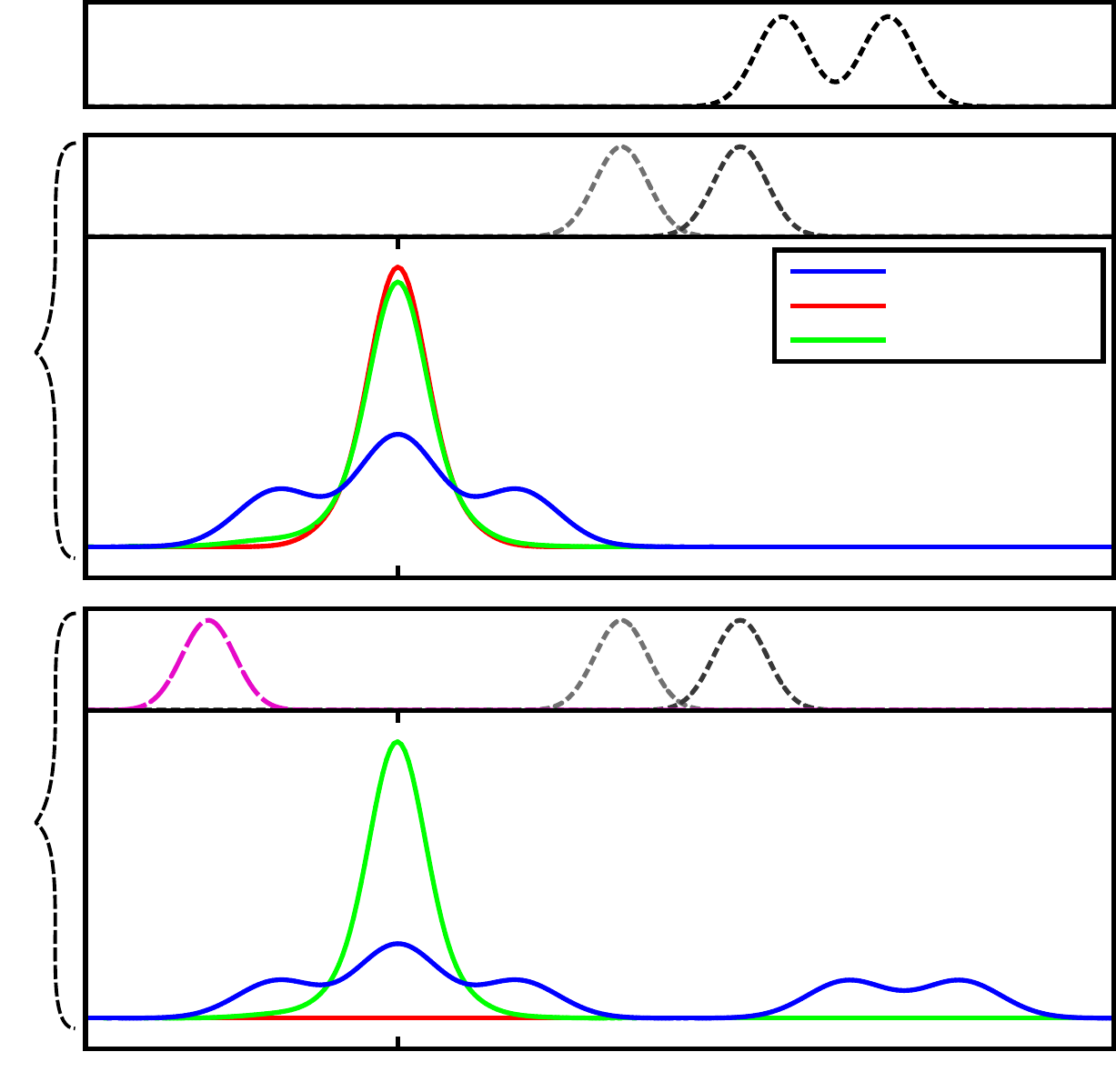}};
  \begin{scope}[x={(img.south east)},y={(img.north west)},every path/.style={draw=black,thick}]
    \draw (0.08,1) node[anchor=north west,text width=27mm,font=\footnotesize,align=left] {previous point set\\(modeled as GMM)};
    \draw (0.08,0.88) node[anchor=north west,text width=27mm,font=\footnotesize,align=left] {current point set\\(2 targets)};
    \draw (1,0.446) node[anchor=north east,text width=27mm,font=\footnotesize,align=right] {current point set\\(3 targets)};
    \draw[line width=0.3mm,<-] (0.19,0.40) -- ++(-0.03,-0.08) node[anchor=north,font=\footnotesize] {an outlier};
    \draw[line width=0.3mm,red,<-] (0.32,0.075) -- ++(0.12,0.10) node[anchor=west,font=\footnotesize] {likelihood fails};
    \draw (0.36,0.002) node[inner sep=0,anchor=south,font=\itshape\footnotesize] {ground truth};
    \draw (0.00,0.25) node[inner sep=0,anchor=north,text width=27mm,font=\bfseries\footnotesize,align=center,rotate=90] {with outlier};
    \draw (0.00,0.68) node[inner sep=0,anchor=north,text width=27mm,font=\bfseries\footnotesize,align=center,rotate=90] {no outlier};
    \draw (0.8,0.75) node[inner sep=0,anchor=west,font=\footnotesize] {Summing};
    \draw (0.8,0.72) node[inner sep=0,anchor=west,font=\footnotesize] {Likelihood};
    \draw (0.8,0.688) node[inner sep=0,anchor=west,font=\footnotesize] {Our approach};
  \end{scope}
\end{tikzpicture}
\caption{A simple 1D cost scenario shows the impact of an outlier on the likelihood and summing approaches compared to our method.
  It reveals a robust behavior for the summing approach against outliers and shows the appearance of multiple local maxima.
  In contrast, our approach shows a robust behavior against outliers while remaining almost identical to the likelihood solution in the non-outlier case.}
\label{fig:1dScenario}
\end{figure}


Recently, radar has been used in different automotive applications, such as object tracking or ego-motion estimation.
While the latter is the focus of our work, we aim for a solution to be independently used as part of a sensor fusion or SLAM framework -- like a virtual odometry sensor.
Specifically, we use the term ego-motion to describe the relative transformation of the ego-vehicle between two consecutive measurements or scans.
The used pipeline for estimating the ego-motion typically varies based on the measurement properties of the sensor -- e.g., is the environment sparsely or densely represented.
For sensors with a sparse measurement, point set registration (PSR) is part of the motion-estimation problem.
PSR aims to find the best alignment between two point sets.
Various algorithms emerged over time, but we still see a gap when it comes to the robustness and credibility of these algorithms.
Particularly, if we think of integrating motion-estimation results into a loosely-coupled sensor fusion framework, we need a fully probabilistic approach to be consistent in the end.

We tackle the PSR problem as part of the motion estimation probabilistically to cope with the challenges regarding outliers in the automotive radar measurements.
By modeling one of the two consecutive point sets as a Gaussian Mixture Model (GMM), we can omit the often needed preprocessing step for finding corresponding points.
\autoref{fig:1dScenario} shows a simple 1D PSR problem, with the reference point set modeled as GMM and two scenarios for a consecutive (current) point set, with and without an outlier.
While using GMMs is not new in this context, we optimized the problem in a feasible time without violating the probability's assumptions in the objective function -- this utilizes our earlier work {\cite{Pfeifer21}} and will be explained further in \autoref{ProbME}.
In consequence, we get a consistent estimation result, meaning that the calculated covariance represents the estimation's true uncertainty -- as can be seen in \autoref{Eval}.
The \hyperref[relWork]{following Section} reviews the existing ego-motion estimation approaches.


In summary, our main contributions are as follows:
\begin{inparaenum}
    \item we model and solve the objective function for an automotive ego-motion radar estimation in a thorough probabilistic way without explicit point-to-point correspondences and
    \item we investigate and compare the consistency of our approach to the often used summing approximation of this problem.
\end{inparaenum}
Furthermore, we provide an exhaustive credibility evaluation on different scenarios and release our implementation in the spirit of transparency and benchmarking. 

%% file: inc/prior.tex

\section{Related Work}
\label{relWork}

Radar sensors got miniaturized over the last years and became a standard sensor in the automotive domain.
These sensors typically provide a limited number of targets in each measurement, like the ARS 408-21 from Continental used in the nuScenes datasets \cite{Caesar20} or the General Purpose Radar from Bosch used in our experiments.
Instead of these commonly used sensors, the authors in \cite{Barnes20} or \cite{Kim20} used the Navtech CTS350-X, which returns 1D power readings while rotating \SI{360}{\degree} resulting in a dense 2D intensity image. 
In consequence, they need an additional keypoint extraction step.
Recently, low-cost solutions like the AWR1843 from Texas Instruments prove to be applicable even to indoor ego-motion estimation \cite{Almalioglu21}.
They provide 3D target measurements through a special transmitter antennas setup to estimate the detection's elevation. 
Within this paper, we use the sparse automotive type and concentrate on its specific properties.
However, most of the addressed challenges like the high outlier rate also apply to the other sensor variants.


Estimating ego-motion is usually a pipeline starting with raw sensor information. 
It aims to compute a robust and credible motion estimation, while the intermediate steps define the estimation algorithm itself.
Depending on the sensor's measurement information, the estimation pipeline needs to be adapted.
E.g., in \cite{Cen18}, the authors needed a feature detection and feature matching step for the Navtech radar sensor, as the sensor represents its surroundings in a dense 2D intensity image. 
However, as common automotive radars include such a step already, available approaches usually aim to find the best alignment between two sparse point sets, without the feature detection step, like in {\cite{Jian11}} or {\cite{Stoyanov12a}}.


In the following, we try to give a brief overview of used algorithms for the PSR problem in general and in relation to automotive Doppler radar sensors.
Specifically, we focus on pure motion estimation algorithms feasible to integrate into a larger sensor fusion framework, which is credible in the end -- often, this is referred to as \emph{loosely-coupled} integration.
Consequently, we need to stress the algorithms' covariance estimation capabilities.

\emph{Iterative Closest Point} (ICP) is a well-known approach to address the PSR problem.
The authors of \cite{Besl92} introduced the standard ICP approach using an explicit correspondence technique to build the error metric but ignoring possible uncertainty of target measurements within the metric.
Likewise, they did not discuss how to calculate the estimation's uncertainty.
This is done later by \cite{Censi07}, where the author follows the error propagation model to additionally estimate the ICP's uncertainty.
Using the error propagation model is motivated by the missing target noise model within the ICP's error metric.
In addition to that, the authors of \cite{Segal09} provided an approach to also integrate the targets' standard deviation into the error metric.
This is done in two steps: constructing correspondence pairs using the Euclidean distance metric and following a probabilistic framework to build the cost function for the point pairs.

\emph{Normal Distribution Transformation} (NDT) is a probabilistic, cell based, condensed representation of measurements, introduced in \cite{Biber03} for a 2D laser range finder.
Additionally, the authors proposed a scan matching algorithm, applying a point-to-distribution error metric.
In \cite{Magnusson07} the authors extended this approach for three dimensions and achieved a 3D-NDT registration while implementing a distribution-to-distribution error metric.
Both works miss a thorough discussion about the estimation's uncertainty, though.
Later in \cite{Stoyanov12a}, an uncertainty estimation was added by applying the error propagation described in \cite{Censi07}. In addition to that, {\cite{Jian11}} introduced a method to weight the NDT cells based on the sensor's noise.

A full \emph{probabilistic approach} to the PSR problem was introduced in \cite{Jian11}.
The authors used a Gaussian Mixture Model (GMM) to represent not only the model point set but also the current point set.
In consequence, the registration problem lies in aligning two GMMs. 
To solve this, the authors introduced the $\vL_2$ distance as a statistical metric for minimizing the GMMs' alignment error. 
Later, the authors in \cite{Barjenbruch15} applied this approach for an ego-motion estimation based on an automotive Doppler radar.
Additionally, they joint the spatial and Doppler information to suppress the impact of moving targets.
In \cite{Rapp17}, the former approach is combined with the NDT representation, and a method for calculating the estimator's uncertainty is integrated.
For this, they follow the error propagation model of \cite{Censi07}.

For completeness, we also refer to the idea of registration-free approaches where the vehicle's motion is estimated based only on the current radar scan.
In \cite{Kellner13}, the authors used the Doppler information and the targets' azimuth angle to estimate the motion vector (in 2DoF), using a single-track Ackermann constraint.
Then, the authors proposed a second work \cite{Kellner14}, which optimizes a full 2D vehicle motion state (3DoF) using at least two Doppler radar sensors.
In {\cite{Park21}}, the authors extend {\cite{Kellner14}} to achieve a full 3D ego-motion, using two mmWave radars in an orthogonal configuration.
In the same vein, \cite{Monaco20} proposed the RADARODO approach, where it follows the idea of \cite{Kellner13} to estimate the translational motion and the NDT approach for the yaw rate estimation. 
Recently, {\cite{Kramer20}} introduced a radar-inertial sensor platform to estimate a 3D velocity through jointing the linear velocity with yaw rate from an IMU.

\paragraph*{Discussion} 
We are focusing on an uncertainty-aware ego-motion estimation algorithm for sensors returning a very sparse environment representation with high outlier probability, e.g., an automotive Doppler radar.
Having this in mind, ICP may come up as a possible way of solving at least the PSR problem in this task, but mainly because of the high amount of outliers, it is not feasible.
With the work using the NDT representation, we mentioned another possible way of solving the PSR problem.
Even though it is robust against outliers, it is more suitable for sensors returning a dense environment representation.
Building the NDT with a very limited number of points will be difficult and insufficient.
Furthermore, most of the approaches use a summing-based likelihood approximation to be robust against outliers.
This may not be obvious at first glance, but even if the correct measurement noise is used, the likelihood approximation corrupts the estimator's internal uncertainty representation.
Thus, most of these approaches need to utilize the error propagation model to return an uncertainty estimation.
In contrast, we propose a fully probabilistic framework based on correct likelihood joints with feasible computation effort by leveraging our earlier work {\cite{Pfeifer21}}.
Moreover, we provide a way to make the likelihood robust against the impact of outliers, as well as get credible results of the estimator's uncertainty.

%% file: inc/algorithm.tex

\section{Probabilistic Motion Estimation}
\label{ProbME}

In the following, we derive our full likelihood implementation and clarify the differences to the sum approximation method {\cite{Stoyanov12a}}.
The estimation process is additionally visualized in \autoref{fig:1dScenario} as a simple 1D scenario to give an intuitive summary of the algorithms.
As part of the motion estimation, PSR aims to find the best transformation to align two point sets.
In the automotive domain, PSR aligns the current or moving point set $\cM$ with the previous or model point set $\cF$, to find the best estimate for the state $\vx =  \left[ \vx^{\text{tran}},  \vx^{\text{rot}} \right] \T$.
We assume, that each point set consists of $K$ targets, given in polar space $\check\vm_k = \left[r_k,\theta_k\right]\T$, where $r$ is the radius and $\theta$ the azimuth angle. 
Further, we assume information about the targets' standard deviation with $\sigma_{r_k}$ and $\sigma_{\theta_k}$ for both dimensions.
In Cartesian coordinates, we speak of $\vm_k$ for the target's position within the sensor frame and $\vSigma_k$ for its propagated covariance.
For including the velocity measurements into the process, we additionally need the targets' Doppler information with $v_k$ and $\sigma_{v_k}$.


Following \cite{Jian11} or \cite{Stoyanov12a}, the measurement's point set could also be represented as multivariate GMM with $K$ components. 
If we do this for the model point set with $K=|\cF|$ targets, we get
\begin{equation}
	\label{eq:step1_Gmm}
	P(\vm \mid \cF) = \sum_{j=1}^{|\cF|} w \cdot \normal{\vm}{\vm_j, \vSigma_j} \qquad \text{with } w = \frac{1}{\left| \cF \right|}.
\end{equation}
This represents a measurement state probability of an arbitrary point $\vm$ within the model coordinate frame. 
The GMM components $\normal{\vm}{\vm_j, \vSigma_j}$ represent the target measurements. 
A simple example of two 1D targets, combined into a GMM, is shown in \autoref{fig:1dScenario} titled \emph{previous point set}.


Thinking of the PSR problem, we need an objective function to be optimized. 
As the objective consists of aligning the two scans, we need a distance metric, which accounts for measurement uncertainties in both consecutive point sets.
Meaning a distribution-to-distribution error metric -- often only point-to-distribution is used.
A possible distance metric, which has been used and derived in \cite{Jian11} or \cite{Stoyanov12a}, is the $\vL_2$ metric between two distributions $\normal{\vx}{\vmu_1,\vSigma_1}$ and $\normal{\vx}{\vmu_2,\vSigma_2}$ with
\begin{equation}
	\label{eq:step2_l2}
	D_{L_{2}} = \normal{0}{\vmu_1-\vmu_2,\vSigma_1 + \vSigma_2}.
\end{equation}


By using the GMM of the previous scan \autoref{eq:step1_Gmm} and the distance metric \autoref{eq:step2_l2}, we can formulate the likelihood for a target $\vm_i$ out of $\cM$ to be present in the previous measurement $\cF$ as
\begin{equation}
	\label{equ:gmmdist}
	P\left(\vm_i \mid \cF, \vx\right) \propto \sum_{j=1}^{|\cF|} s_{j} \exp \left(-\frac{1}{2}\left\| \vr_{i}-\vmu_j \right\|_{\vSigma_{j}+\tilde{\vSigma}_{i}}^{2}\right).
\end{equation}
Now, we have an additional condition in the likelihood, which is the state $\vx$.
It is used for the residual function $\vr_i$ (please note the difference to the scalar $r$ -- the radius in polar space of a target) representing the transformation of point $\vm_i$ into the model frame:
\begin{align}
    \vr_i (\vx) = \vR(\vx^{\text{rot}}) \vm_i + \vx^{\text{tran}}.
\end{align}
Please also note that we use $\vmu_j$ now for a GMM component's mean instead of $\vm_j$ to stress that it is not part of a residual function but the noise model. 
Since we transform the measurement of target $\vm_i$, we also have to propagate its covariance matrix by $\tilde\vSigma_i = \vR \vSigma_i \vR\T$ using the current estimate's rotation matrix $\vR(\vx^{\text{rot}})$.
Combined with $\vSigma_j$, the covariance matrix of GMM component $j$, represents the component's noise as of \autoref{eq:step2_l2}.
The weight and a normalizing component is represented by
\begin{equation}
	s_j = w \det \left( \vSigma_{j}+\tilde{\vSigma}_{i} \right)^{-\frac{1}{2}} \qquad \text{with } w = \frac{1}{\left| \cF \right|}.
\end{equation}

\paragraph*{Objective Function}
\label{ObjFun}
Optimizing the vehicle's motion based on the stated PSR problem, is the process of maximizing
\begin{equation}
	\PP{\vx}{\cM,\cF} \propto \PP{\cM}{\cF,\vx},
\end{equation}
where $\PP{\cM}{\cF,\vx}$ is the joint model of all targets' measurement probabilities $\PP{\vm_i}{\cF,\vx}$ -- or expressed in words: how likely is it, that the set $\cM$ is been generated out of $\cF$ with transformation $\vx$.
At this point, we have to differ between the approximation via summing \autoref{eq:cost_Summing} e.g. proposed in \cite{Stoyanov12a} and the natural way of using the full likelihood \autoref{eq:cost_likelihood}.
\begin{align}
	\label{eq:cost_Summing}
	\PP{\cM}{\cF,\vx} &\sim \sum_{i=1}^{|\cM|} \PP{\vm_i}{\cF,\vx}\\
	\label{eq:cost_likelihood}
	\PP{\cM}{\cF,\vx} &= \prod_{i=1}^{|\cM|} \PP{\vm_i}{\cF,\vx}
\end{align}
Again, via the above measurement likelihood, both approaches aim to find the best transformation ($\vx$), maximizing the likelihood for all current targets to have the best correspondence with the model GMM distribution. 


The summing approach (SA) is commonly used in the literature, as it shows a robust behavior against the impact of outliers, whereas the \emph{pure} likelihood approach most likely will fail because of them, due to the mathematical multiplication, see \autoref{fig:1dScenario}. 
However, because of the approximation via summing, the SA induces multiple local maxima, which makes it sensitive to a good initial guess in the optimization step.
Moreover, the covariance estimation is problematic, which will be explained later on.
On the other hand, we have the drawbacks of outlier integration in the likelihood approach, which will be addressed now.


\paragraph*{Cost Function}
For efficient optimization in the context of Gaussian distributed measurements, least-squares optimization is the obvious choice. 
Therefore, the searched maximum likelihood solution
\begin{align}
	\label{eq:objective}
	\hat\vx &= \argmax_\vx \PP{\vx}{\cM,\cF},
\end{align}
of the objective function is transformed into a minimization problem 
\begin{align}
	\label{eq:minimization}
	\hat\vx &= \argmin_\vx -\log \PP{\vx}{\cM,\cF},
\end{align}
using the negative log-likelihood -- now we speak of the cost function. 
In combination with only normal distributed constraints, the cost function simplifies into the common least squares formulation.
However, we have a GMM as the noise model for the model point set.
Therefore, we follow the robust least squares approach in \cite{Pfeifer21}, converting the GMM \autoref{equ:gmmdist} into a loss function $\vrho$ resulting in the following least squares problem:
\begin{equation}
	\hat\vx = \argmin_\vx \sum^{\left| \cM \right|}_{i=1} \frac{1}{2} \smd{\vrho \left( \vr_i \left( \vx \right) \right)}{}.
\end{equation}
For the GMM's loss function implementation, we use the proposed Max-Sum-Mixture (MSM).
This way, we get an optimization problem to be solved in feasible processing time -- please see \cite{Pfeifer21} for a more complete elaboration.

\paragraph*{State definition}
Till now, we did not restrict the estimated transformation $\vx$ in its DoF.
While in general, the presented PSR methods would be applicable to full 3D motion estimation, we can restrict it in our automotive examples to 2D.
Further, we assume a nonholonomic car-like vehicle, reducing our estimated state to 2DoF with $\vx$ having only a longitudinal translation and an angle of rotation.
If a non-negligible sideslip is expected, the method can easily be extended to 3DoF.


\paragraph*{Outlier Model}
In contrast to the SA, which has a natural robustness against outliers, the likelihood approach needs to model them explicitly -- as can be seen in our simple example in \autoref{fig:1dScenario}. 
Robust cost functions or M-estimators like presented in \cite[Appendix.~6]{Hartley03} are one possibility. 
As we already use a loss function implementing a GMM, the natural way is to model the outlier component directly into it.
Therefore, we can approximate the uniform distribution over the measurement space of the radar sensor with additional GMM components and merge the inlier and outlier distribution for the previous scan $\mathcal{F}$.


For the outlier distribution, we approximate a model based on the radar's measurement space as Gaussians in the form of a conic ray, like it is described in \cite[Ch.~6.4.1]{Sola07}. 
There, two parameters $\alpha$ and $\beta$ regulate the number and width of the components generated and two additional parameters $s_{min}$ and $s_{max}$ to limit the range accordingly.
Following the equations, we get range parameters for a number of $N_g$ components.
Transforming it into polar space, we need the additional parameter $\sigma_\theta$ for the angle uncertainty, which is specified according to the sensor's FoV.
Then, we can transform the components into Cartesian space getting $x_k$ and the two by two covariance $\vSigma_k$ to get the GMM
\begin{equation}
	\label{equ:outl}
	\PP{\vm_{\text{o}}} = \frac{1}{\sum w_k} \cdot \sum_{k=1}^{N_g} w_k \cdot \normal{\vm}{\twovector{x_k}{0},\vSigma_k}
\end{equation}
with its components weighted by $w_k = \sqrt{\det \vSigma_k}$.

Finally, this new GMM \autoref{equ:outl} has to be combined with the model GMM \autoref{equ:gmmdist} for each current measurement:
\begin{equation*}
	\resizebox{0.95\hsize}{!}{%
	$\PP{\vm_i}{\cF,\vx,\vm_\text{o}} = w_\text{o} \PP{\vm_\text{o}} + \left( 1- w_\text{o}\right) \PP{\vm_i}{\cF, \vx}$%
	}
\end{equation*}
where it introduces a tuning parameter $w_\text{o}$, which represents the expected amount of outliers -- it weights both sets of GMM components accordingly.

\paragraph*{Doppler Integration}
Till now, we addressed ego-motion estimation as an application for PSR utilizing range sensors like lidar and radar.
However, apart from spatial information, the automotive radar provides Doppler measurements.
In \cite[equ.~14]{Barjenbruch15}, these are jointly used with the correspondence likelihood to suppress non-stationary objects.
We followed this approach in our SA implementation, as can be seen in \autoref{eq:PWithDoppler_summing}.
However, as the Doppler information is a nearly independent measurement, we could also model it like an extra sensor within the objective function, which we did in our likelihood approach, see \autoref{eq:PWithDoppler_likelihood}.
\begin{align}
	\label{eq:PWithDoppler_summing}
	\PP{\cM}{\cF,\vx} &\sim \sum_{i=1}^{|\cM|}  \PP{\vm_i}{\cF,\vx} \PP{v_i}{\vx, \theta_i,\Delta t} \\
	\label{eq:PWithDoppler_likelihood}
	\PP{\cM}{\cF,\vx} &= \prod_{i=1}^{|\cM|} \PP{\vm_i}{\cF,\vx} \PP{v_i}{\vx, \theta_i,\Delta t}
\end{align}

With the Doppler information, we get a good measurement of the vehicle's longitudinal translation over time.
Even the information about the relative rotation may be improved if the sensor is not located in the vehicle's center of rotation.
A measurement model for the expected Doppler velocity of each target can be found in \cite{Barjenbruch15}. 
As we work with relative transformation in our state definition $\vx$ instead of velocities, we use the proposed model with small modifications, as stated in \autoref{eq:PDoppler}.
\begin{equation}
	\label{eq:PDoppler}
	\PP{v_i}{\vx, \theta_i,\Delta t} \propto \exp \left(-\frac{1}{2} \smd{\hat{u}_i - v_i \cdot \Delta t}{\hat{\sigma}_{\theta_i}^{2}+\sigma_{u_i}^{2}} \right)
\end{equation}
Here, $\hat u_i$ is the measurement model \autoref{eq:dopplerModel} for the expected Doppler velocity, expressed as relative translation.
In comparison to \cite{Barjenbruch15}, we omit the lateral translation, as we use only 2DoF in combination with Doppler information in our implementation.
Further, we use the same notation for the sensor offset relative to the car's center of rotation with $x_S$, $y_S$, and $\alpha_S$.
With $\hat\sigma_{\theta_i}$, we account for the measured and uncertain azimuth angle of target $i$ -- see \autoref{eq:SigmaDopplerModel}.
\begin{align}
	\label{eq:dopplerModel}
	\hat u_i &= \begin{bmatrix}
			\vx^\text{rot} \cdot y_S - \vx^\text{tran}\\
			- \vx^\text{rot} \cdot x_S
		\end{bmatrix}\T \cdot
		\begin{bmatrix}
			\cos(\theta_i + \alpha_S)\\
			\sin(\theta_i + \alpha_S)
		\end{bmatrix}\\
	\label{eq:SigmaDopplerModel}
	\hat\sigma_{\theta_i}^2 &= \left( \tfrac{\partial \hat u_i}{\partial \theta_i} \sigma_{\theta_i} \right)^2 
\end{align}
In consequence of our state definition, we do not use the pure Doppler measurement in \autoref{eq:PDoppler}, but transform it by using the time difference between the measurements with $\Delta t$.
This measurement may also be uncertain, why we need to consider it in our measurement variance $\sigma_{u_i}^2$, see \autoref{eq:SigmaDopplerMeasurement}.
{
\newcommand{\DzDv}{\pfrac{u_i}{v_i}}
\newcommand{\DzDt}{\pfrac{u_i}{\Delta t}}
\newcommand{\JMeasurement}{\begin{bmatrix} \DzDv & \DzDt \end{bmatrix}}
\begin{align}
	u_i &= v_i \cdot \Delta t \\
	\label{eq:SigmaDopplerMeasurement}
	\sigma_{u_i}^2 &= \JMeasurement \cdot \diag (\sigma_{v_i}^2, \sigma_{\Delta t}^2) \cdot \JMeasurement\T
\end{align}
}

\paragraph*{Covariance Estimation}
Unfortunately, the importance of a good covariance estimate is often neglected, and the evaluation focus of new registration approaches lies mostly on the mean error.
Aiming at more complex sensor fusion with several sensors and possible pre-processing steps to reduce the problem's size, we need to stress the consistency of a pre-processing step like the present ego-motion estimation.
An algorithm with a good mean error but a bad covariance estimate will probably lead to a bad overall performance compared to the one with less quality in the mean error but consistent covariances.
In work like \cite{Censi07} or \cite{Stoyanov12a}, the covariance is calculated in a post-registration step, as the measurements' noise models are not taken into account within the registration step. 
In contrast to this, we have all necessary noise models already included within the optimization problem.
Accordingly, we can get the estimate's covariance by inverting the Hessian, which is a byproduct of the optimization\footnote{To be exact: typically, this is an approximated Hessian based on the squared Jacobian.} and represents the cost function's shape around the optimized value.
Further, as we are minimizing a negative log-likelihood, we can also refer to the Hessian as Fisher Information Matrix (FIM), indicating how much information about the estimated state is given by the observations.

\paragraph*{Implementation}
The presented likelihood approach is implemented within GTSAM and used through the accompanied Matlab wrapper.
For implementing the objective \autoref{eq:objective} based on \autoref{eq:PWithDoppler_likelihood}, one variable vector representing a translation and rotation ist used. 
Each targets' position within the current measurement is used in a unary factor, connected to the variable vector.
The previous measurement and outlier distribution is represented within a special noise model for GMMs and added to each of these factors. 
Additionally, the same amount of unary factors is added for the Doppler measurements.
For additional robustness against local minima, we also used the heuristic concept of covariance scaling (e.g., mentioned by \cite{Jian11}) to get better initialization points, meaning that we first run a maximum of 5 iteration steps with scaled covariances (by factor 5), before the correct ones are used.

For the SA, a pure Matlab implementation leveraging the nonlinear programming solver is used.

%% file: inc/eval.tex

\section{Evaluation}
\label{Eval}
We start our evaluation with a common PSR problem, motivated by the idea to compare against a standard ICP implementation using error propagation for the covariance estimate.
In the following experiment, we extended this Monte Carlo simulation with regard to representing automotive radar sensor measurements.
With the perfectly-known measurements' uncertainty in both simulation scenarios, a consistency analysis can be done.
Finally, real-world data gathered with a mobile robot and a car is evaluated to show the applicability to real-world scenarios, whereas the automotive scenario is based on the publically available nuScenes dataset. 
According to our main objective of a loosely-coupled algorithm, all experiments evaluate only the scan-to-scan performance with no previous knowledge, i.e., zero initialization.

\begin{table}[tb]
	\centering
	\caption{Parameters for the Simulation Experiments}
	\label{tab:parameters}
	\setlength\tabcolsep{4pt}
	\sisetup{round-mode = off,separate-uncertainty = true}
	\begin{tabular}{@{}m{2.9cm}cc@{}}
		\toprule
		\textbf{Experiment}         & \textbf{PSR} & \textbf{Radar} \\ \midrule
		Num. of Configurations      & 100 & 50 \\%
		\arrayrulecolor{lightgray}\midrule 
		Num. of Landmarks      & 20 (+16) & 20 (+16) \\%
		\arrayrulecolor{lightgray}\midrule 
		Landmark\newline Generation & 
		$\begin{aligned} &\uniform{r}{\SI{10+-5}{\metre}}      \\ &\uniform{\theta}{\pm\pi}\end{aligned}$ & 
		$\begin{aligned} &\uniform{r}{\SI{20+-18}{\metre}}      \\ &\uniform{\theta}{\SI{\pm 55}{\degree}}\end{aligned}$ \\%
		\midrule
		Clustering (size)           & \SI{40}{\percent} (\num{3}) & \SI{40}{\percent} (\num{3}) \\ 
		Cluster spread              & $\normal{x,y}{0,0.1^2}$ & $\normal{x,y}{0,0.1^2}$ \\%
		\arrayrulecolor{black}\midrule
		Runs per Configuration      & 1000 & 500 \\%
		\arrayrulecolor{lightgray}\midrule
		Transformation\newline sampling &
		$\begin{aligned} &\uniform{x,y}{\SI{+-0.25}{\metre}}   \\ &\uniform{\alpha}{\ang{+-15}}\end{aligned}$ & 
		$\begin{aligned} &\uniform{x}{\SI{+-0.25}{\metre}}   \\ &\uniform{\alpha}{\ang{+-15}}\end{aligned}$\\%
		\arrayrulecolor{lightgray}\midrule
		Measurement noise           &
		$\begin{aligned} &\normal{r}{0,\left(\SI{0.2}{\metre}\right)^2}\\ &\normal{\theta}{0,\left(\ang{3}\right)^2}\end{aligned}$ &
		$\begin{aligned} &\normal{r}{0,\left(\SI{0.2}{\metre}\right)^2}\\ &\normal{\theta}{0,\left(\ang{3}\right)^2}\end{aligned}$ \\%
		\arrayrulecolor{black}\bottomrule
	\end{tabular}
\end{table}

\subsection{Simulated Point Set Registration}
In PSR, we try to estimate a transformation aligning two sets of points. 
Regarding the radar-based motion estimation at hand, this is referred to as the spatial registration and can be evaluated separately.
Accordingly, we generate multiple 2D PSR problems with 3DoF\footnote{Note: Only for the PSR experiment, we ignore the nonholonomic constraint and use a full 2D state for better comparison to PSR problems in general.} evaluating the algorithms' accuracy and uncertainty output (i.e., credibility).
As already mentioned, in this scenario, we also compare against the standard ICP.
The implementation is based on the well-known PCL-Library \cite{Rusu11}, combined with covariance estimation based on the accompanied implementation of \cite{Prakhya15}.

We simulated \num{100} different landmark configurations and \num{1000} different transformations. 
All transformations are applied to each landmark configuration to generate the two point sets, which results in \SI{100}{\k} experiments.
Gaussian measurement noise is applied independently to each of the \num{20} landmarks within the \SI{200}{\k} point sets in polar space.
The landmark configurations themselves are sampled from two uniform distributions in polar space ($[r, \theta]$) as also specified in \autoref{tab:parameters}.

As part of this synthetic PSR scenario, we run the \SI{100}{\k} experiments again with clustered landmarks -- clustered landmarks often appeared in our radar measurements.
Specifically, we randomly select \SI{40}{\percent} out of the \num{20} landmarks to duplicate them two times using the spread parameters given in \autoref{tab:parameters}.
As a result, we now have 12 unmodified landmarks and 8 clusters of three landmarks -- indicated in \autoref{tab:parameters} with $(+16)$. 

\subsection{Simulated Radar}

For the synthetic radar dataset, the PSR experiment is repeated but with small modifications. 
Considering now a sensor with limited FoV, we generate the landmarks only in front of the sensor. 
Naturally, the limited FoV in combination with the applied transformations for the second point set fosters an outlier appearance. 
In consequence, the number of landmarks, shown in \autoref{tab:parameters}, is only the number of created landmarks.
However, the number of corresponding landmarks for estimating the motion estimation is less. 
The ratio between corresponding and non-corresponding landmarks is shown in \autoref{fig:LmOutliers}.
Again, we also simulate a problem with clustered landmarks, and additionally, a Doppler velocity is simulated for each of the landmarks with a standard deviation of \SI{0.3}{\metre\per\second}.

\newcommand{\splitcell}[1]{\begin{tabular}[c]{@{}c@{}}#1\end{tabular}}
\newcommand{\nanf}[1]{{\cellcolor[gray]{0.8}} #1}
\begin{table}[tbp]
	\centering
	\caption{Summarized Results}
	\label{tab:result}
	\setlength{\tabcolsep}{4pt}
	\sisetup{round-mode = places,
	         round-precision = 2,
			 detect-weight = true,
			 detect-inline-weight = math}
	\begin{tabular}{@{}
					  c
					  r
					  S[table-format=1.3, round-precision = 3]
					  S[table-format=1.2]
					  S[table-format=1.2]
					  S[table-format = 2.1,round-precision = 1]
					  S[table-format = 2.1,round-precision = 1,table-number-alignment = center]
					  @{}}
		\toprule
		&
		\multicolumn{1}{c}{Alg.} & 
		{\splitcell{RMSE\\ \hspace{0pt}[\si{\metre}]}} & 
		{\splitcell{RMSE\\ \hspace{0pt}[\si{\degree}]}} & 
		{ANEES} & 
		{\splitcell{Average\\ Iterations}} &
		{\splitcell{Average\\ Runtime [\si{\milli\second}]}} \\ \midrule
	    \multirow{4}{*}{\begin{tabular}[c]{@{}c@{}}PSR\end{tabular}}%
			& ICP &  0.20005 & 1.30159 & 1.30588 & 3.50000 &  0.28875 \\ 
			& MSM & 0.12094 & 0.98866 & 1.07123 & 4.20000 &  1.97912 \\ 
			& SA &  0.26567 & 10.57957 & \color{red!30} 1.56553 & 13.60000 &  26.91891 \\ 
			\midrule
		\multirow{4}{*}{\begin{tabular}[c]{@{}c@{}}PSR-C\end{tabular}}%
			& ICP &  0.15649 & 1.07285 & 1.51039 & 5.90000 &  0.33198 \\ 
			& MSM &  0.09675 & 0.76575 & 1.21239 & 5.50000 &  3.75811 \\ 
			& SA &  0.20639 & 8.22075 & \color{red!30} 0.72385 & 13.60000 &  74.38604 \\ 
			\midrule
		\multirow{4}{*}{\begin{tabular}[c]{@{}c@{}}Sim\end{tabular}}%
			& MSM &  0.05199 & 0.34893 & 0.98165 & 6.40000 &  2.83790 \\ 
			& SA &  0.26557 & 8.57238 & \color{red!30}23.64383 & 9.80000 &  10.65222 \\ 
			& MSM-D &  0.01675 & 0.34650 & 1.01956 & 6.10000 &  5.68517 \\ 
			& SA-D &  0.04919 & 8.54290 & \color{red!30}21.80811 & 8.10000 &  9.62264 \\ 
			\midrule
		\multirow{4}{*}{\begin{tabular}[c]{@{}c@{}}Sim-C\end{tabular}}%
			& MSM &  0.04410 & 0.28582 & 1.25661 & 7.50000 &  4.73666 \\ 
			& SA &  0.25599 & 8.32964 & \color{red!30}15.72037 & 10.00000 &  26.98151 \\ 
			& MSM-D &  0.01226 & 0.27848 & 1.08559 & 6.90000 &  9.79112 \\ 
			& SA-D &  0.04007 & 8.30944 & \color{red!30}16.89886 & 8.00000 &  23.48146 \\ 
			\midrule
		\multirow{4}{*}{\begin{tabular}[c]{@{}c@{}}Robo\end{tabular}}%
			& MSM &  0.10657 & 1.11717 & 2.52711 & 8.20000 &  2.95180 \\ 
			& SA &  0.08451 & 1.35821 & \color{red!30}0.16009 & 7.50000 &  11.01560 \\ 
			& MSM-D &  0.01096 & 1.11321 & 1.68344 & 5.90000 &  6.04314 \\ 
			& SA-D &  0.01704 & 1.36538 & \color{red!30}0.20375 & 6.10000 &  9.36364 \\ 
			\midrule
		\multirow{4}{*}{\begin{tabular}[c]{@{}c@{}}NuSc\end{tabular}}%
			& MSM &  0.16306 & 0.15368 & 1.60820 & 11.10000 &  23.23852 \\ 
			& SA &  0.09149 & 0.12678 & \color{red!30}0.00463 & 11.20000 &  242.85425 \\ 
			& MSM-D &  0.01158 & 0.10496 & 3.35996 & 6.20000 &  33.96818 \\ 
			& SA-D &  0.22621 & 5.30589 & \color{red!30}2.32828 & 7.20000 &  225.04392 \\ 
			\midrule
		\multirow{2}{*}{\begin{tabular}[c]{@{}c@{}}NuScT\end{tabular}}%
			& MSM &  0.16764 & 0.15223 & 1.63448 & 10.90000 &  22.67567  \\ 
			& MSM-D &  0.01216 & 0.09555 & 4.49242 & 6.10000 &  34.34684 \\ 
			\bottomrule
	\end{tabular}
\end{table}

\subsection{Mobile Robot Dataset}
Our main motivation for using the radar sensor in a loosely-coupled way for motion estimation emerged from developing a robust sensor fusion algorithm for two mobile robots, described in \cite{Lange16}. 
Lately, we extended the robots' sensor setup by an automotive radar\footnote{The specific type is a General Purpose Radar V1.0 from Bosch.} sensor.
The sensor is mounted near to the center of the robot and detects up to 48 targets with \SI{10}{\hertz}. 
It is interfaced via CAN-bus and measures the horizontal angle, distance, Doppler velocity, and additional properties for each target.
Among the additional properties, a standard deviation for the measurements is given.

For evaluation of our algorithm, we recorded a dataset at the TU Chemnitz campus. The area is mostly surrounded by buildings with a small street and about \SI{60}{\percent} grass-covered with some trees.
Unfortunately, no exact ground truth is available. 
So, we used batch optimization over several sensor information originating from wheel odometry, IMU, and an external UWB distance-measuring system instead.
Based on visual verification of the calculated trajectory with the satellite images, we claim to have a remaining error of less than \SI{0.5}{\metre}.
We additionally compared our trajectory to a solution calculated using the Cartographer framework \cite{Hess16} based on measurements recorded with an onboard Velodyne VLP-16, resulting in negligible differences.

In contrast to the synthetic scenarios, information about ground truth correspondences between consecutive scans is not available.
However, this is an important evaluation metric, so we use a simple classification scheme for correspondence estimation based on the Bhattacharyya distance \cite{Hennig10}
\begin{align}
	D_\text{Bhat} &= \frac{1}{8} \smd{\vmu_1 - \vmu_2}{\bar\vSigma} + \frac{1}{2} \ln \left( \left| \bar\vSigma \right| \left( \left| \vSigma_1 \right| \left| \vSigma_2 \right| \right)^{-\frac{1}{2}} \right).
\end{align}
Basically, this gives us a similarity rating between two distributions $\normal{\vx}{\vmu_1,\vSigma_1}$ and $\normal{\vx}{\vmu_2,\vSigma_2}$, where smaller values represent higher overlapping. 
For clarification, $\bar\vSigma$ stands for a mean with $\bar\vSigma = 0.5 \cdot \left(\vSigma_1 + \vSigma_2\right)$, $\smd{.}{\vSigma}$ and $\left| . \right|$ is the squared Mahalanobis distance respective the determinant.
Using the vehicle's ground truth, we can transform the current scan and evaluate $D_\text{Bhat}$ for all possible landmark combinations.
Then, we select for each landmark in the current scan one landmark with the smallest distance from the previous.
A landmark pair is then classified as correspondence if the distance is smaller than a threshold.
We defined the empirically determined threshold to be \num{1}. 
All landmarks within the current and the previous scan without correspondence represent the number of outliers or non-corresponding landmarks.
In \autoref{fig:LmOutliers}, the classification result can be seen.
Compared to the simulation scenario, the amount of non-corresponding targets is increased.
Further, some motion estimates will be based on a very small number of correspondences.

\subsection{nuScenes}
With the nuScenes dataset \cite{Caesar20}, we chose a publicly available dataset for further evaluation and the possibility of benchmarking against our solution.
The dataset provides ground truth and automotive radar data\footnote{The used type is an ARS 408-21 from Continental.} for \num{850} driving scenes with changing complexity each of \SI{20}{\second} in length.
Out of the five radar sensors the car is equipped with, we use only the front radar, which is captured with \SI{13}{\hertz} and a maximum of 125 targets.
Each measurement can be read as a point cloud with additional information like Doppler velocity, standard deviation, or quality classification for each target -- e.g. regarding a validity or ambiguity state.
Each scene also includes the extrinsic calibration describing the sensor's position and angle within the car's body frame.
In contrast to our other experiments, the sensor's lever arm increases the angular accuracy when using doppler information.

Our evaluation is done for the \emph{mini} dataset and the first \num{500} of the \num{700} \emph{trainval} training scenes.
For each dataset, we consider all scenes together as a single experiment, resulting in \num{2600} respective \SI{130}{\k} estimation problems.
As the scenes are very different in their properties, it is also interesting to look at each scene separately, which we do for the \emph{trainval} dataset.
For the \emph{mini} dataset, the outlier statistics are shown again in \autoref{fig:LmOutliers}, calculated in the same way as for the mobile robot dataset.
Compared to the other datasets, we generally have more landmarks but a similar outlier to correspondence ratio like in the mobile robot dataset. 

Side notes: 
\begin{inparaenum}
	\item To remain consistent with our algorithms, we convert the dataset's radar sensor information regarding the given Doppler information from Cartesian coordinates to an absolute Doppler velocity pointing towards the sensor, including the given noise. 
	\item As we convert the Doppler into a translation by using the time difference between two consecutive measurements, we need to give it an uncertainty, which is $\sigma = \SI{5}{\milli\second}$. This can be evaluated by the timestamps, which fluctuate around \SI{13}{\hertz}.
\end{inparaenum}

%% file: inc/results.tex

\section{Results}

In the following, we evaluate our results based on two metrics:
\begin{inparaenum} 
    \item the relative angle and position error as RMSE\footnote{The $\text{RMSE}\correspondto(\frac{1}{m}\sum_{k=1}^m \smd{\text{err}_k}{})^{1/2}$ where $\text{err}_k$ corresponds to either the translational or rotational part of the scan-to-scan performance compared to the ground truth.} and 
    \item the average normalized estimation error squared (ANEES).
\end{inparaenum}
For the ANEES, we follow \cite[Chp.~5.4]{Shalom01}, but normalize with the problem's dimension as stated in \cite{li02}.
The later metric gives us insight into the goodness (or credibility) of the estimators' covariance calculation. 
If the estimator is perfectly credible, the ANEES value equals \num{1}. 
However, this holds only under the assumption of a $\chi^2$-distributed NEES -- otherwise, the ANEES is not meaningful.
Additionally, we provide the average runtime per registration, evaluated on a standard AMD Ryzen 7 desktop PC. 

\paragraph*{Simulation}
We start by comparing the results of the simulated experiments.
Because of a bigger chance to reach local minima, the sum approximation has a higher RMSE for both, the translation and rotation, as can be seen in \autoref{tab:result} (PSR, Sim, PSR-C, and Sim-C -- C refers to the clustered dataset).
Of course, the common ICP approach with its closed-form covariance estimation is the fastest solution but does not include the landmarks' covariances within the registration process. 
Accordingly, it performs a bit worse regarding RMSE and consistency.
Generally, we find the ANEES values for all simulation experiments for our likelihood approach reasonably close to one, which means a consistent estimation.
The estimated covariance based on the SA, on the other hand, is not useful, as the SA violates the Gaussian assumption by adding the probabilities instead of jointing them. 
This results in NEES values not following a $\chi^2$-distribution\footnote{A plot is omitted here because of space constraints.}, making the ANEES evaluation inconclusive.
In the simulation with clustered landmarks, more information is introduced into the problem, resulting in a decreased error and increased runtime.
We can also see in the ANEES that the covariance is slightly more overconfident. 
By adding the Doppler information in the simulated radar experiment (marked with -D in \autoref{tab:result}), the translational RMSE gets better -- as we would expect.
Likewise, we see an increase in runtime for the likelihood approach as additional factors are introduced into the optimization problem.

\paragraph*{Real World}
As before, the results are summarized in \autoref{tab:result}, where \emph{Robo} corresponds to the mobile robot dataset.
The combined results of the two considered nuScenes datasets are named \emph{NuSc} and \emph{NuScT} respective for the \emph{mini} and \emph{trainval} variant.
Looking at the mobile robot dataset results, we see basically identical RMSE values for the approximation and the likelihood approach considering the assumed accuracy of the ground truth.
For the nuScenes \emph{mini} dataset however, the RMSE values for the pure spatial estimation show a slightly better performance in advance to the approximation approach, but worse if we include the Doppler information -- again, the approximation approach is more sensitive to local minima.
The \emph{trainval} dataset is only evaluated for the likelihood and confirms the evaluation of the \emph{mini} dataset, as it has nearly identical RMSE errors. 

Regarding the ANEES values, we see that all results show more or less a trend to be overconfident.
The worst result is with about \num{4.5} -- to get an impression, what this means: if we get an ANEES of about \num{4}, we need to multiply the estimated standard deviation by a factor of two to be consistent again.
If we take the uncertainties of the provided data into account, the deviations to the simulated results are explainable.
In real-world scenarios, there is always the question of how good is the sensor's information about the noise.
Unfortunately, this is not provided within the used sensors' datasheets but states a critical parameter for the overall consistency. 
Likewise, there is no information about the algorithm the sensors use for estimating the targets' standard deviation for velocity and position. 
Additionally, further unknowns like the goodness of ground truth, time stability between measurements, or the number of outliers play a role within the evaluation and lead to tunable parameters. 

If we look into the error and consistency statistics over the \num{500} scenes of the \emph{trainval} dataset in \autoref{fig:NuScenesTrainval}, the impact of scene property variation becomes evident.
We see a big spread for the error as well as for the ANEES. 
Further investigation could be done here, in which special situations, the likelihood approach does not perform well. 
Especially in scenes where the car does not move, there is the trend for the ANEES value to become very low.

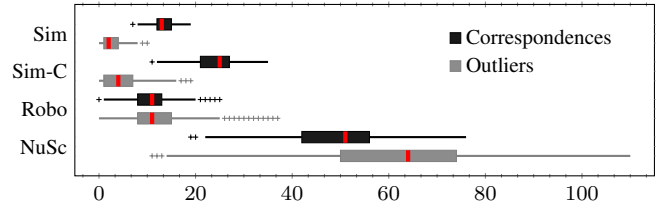
\begin{figure}
	\centering
	\pgfplotsset{every axis x label/.append style={font=\footnotesize, yshift=0.6ex}}
	\pgfplotsset{every tick label/.append style={font=\footnotesize}}
	\setlength\abovecaptionskip{-2pt}
	\input{inc/LmPlot.tex}%
	\caption{The diagram shows the number of corresponding and non-corresponding landmarks for the relevant scenarios. The ratio between inlier and outlier as well as the overall amount of landmarks characterizes the difference between the datasets. Further can be noticed, that especially in the mobile robot dataset, the number of correspondences is in some experiments nearly zero.}
	\label{fig:LmOutliers}
\end{figure}
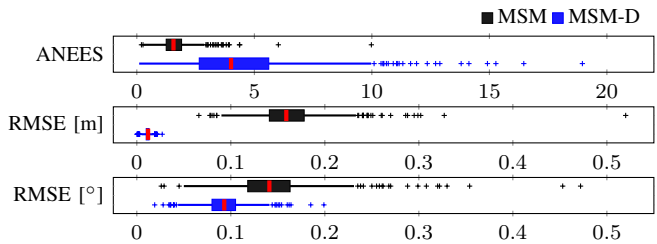
\begin{figure}
	\centering
	\pgfplotsset{every tick label/.append style={font=\footnotesize}}
	\setlength\abovecaptionskip{-2pt}
	\input{inc/NuScenesTrainval.tex}
    \makeatletter
	\caption{To give a deeper insight regarding possible result variations, the plots show evaluation results for the nuScenes \emph{trainval} dataset per scene, whereas the blue boxes correspond to the results using Doppler velocity and the black boxes without. Note, that the axis units for the position RMSE are given in meters and for the angular RMSE in degrees. }
	\label{fig:NuScenesTrainval}
\end{figure}

%% file: inc/LmPlot.tex
\pgfmathsetlengthmacro\MajorTickLength{
      \pgfkeysvalueof{/pgfplots/major tick length} * 0.5
    }
\pgfmathsetlengthmacro\MinorTickLength{
      \pgfkeysvalueof{/pgfplots/minor tick length} * 0.5
    }
\pgfplotsset{
    every axis plot/.append style={fill,fill opacity=0.9},
    every boxplot/.style={mark=x,every mark/.append style={mark size=1.0pt}},
    every boxplot/.append style={
        boxplot/draw/lower whisker/.code={
            \draw[/pgfplots/boxplot/every lower whisker/.try] (boxplot cs:\pgfplotsboxplotvalue{lower whisker})
              -- (boxplot cs:\pgfplotsboxplotvalue{lower quartile});
        },
        boxplot/draw/upper whisker/.code={
            \draw[/pgfplots/boxplot/every upper whisker/.try] (boxplot cs:\pgfplotsboxplotvalue{upper quartile})
              -- (boxplot cs:\pgfplotsboxplotvalue{upper whisker});
              }
        },       
    boxplot/.cd,
        every lower whisker/.style={thick},
        every upper whisker/.style={thick},
        every median/.style={red,ultra thick}
}
\begin{tikzpicture}
\begin{axis}[
  width=0.87\linewidth,
  scale only axis,
  y=0.25cm,
  xmin=-5,xmax=115,
  tick align=center, 
  major tick length=\MajorTickLength,
  minor tick length=\MinorTickLength,
  minor x tick num=5,
  y dir=reverse,
  ytick=\empty,
  yticklabel=\empty, 
  extra y ticks={1,1.5,2,3,3.5,4,5,5.5,6,7,7.5,8},
  extra y tick labels={,Sim,,,Sim-C,,,Robo,,,NuSc,},
  extra y tick style={major tick length=0pt},
  node font=\footnotesize,
  legend style={font=\footnotesize,draw=none,fill=none}, 
  legend cell align=left,
  every axis legend/.append style={at={(0.95,0.92)},anchor=north east}, 
  legend entries={Correspondences,Outliers}
]
\addlegendimage{mark=square*,color=black,only marks, mark size=2pt}
\addlegendimage{mark=square*,color=black!50,only marks, mark size=2pt}
\addplot [
 boxplot prepared={
 lower whisker=8.000,lower quartile=12.000,
 median=13.000,
 upper quartile=15.000, upper whisker=19.000, box extend=0.6
}, mark=+, color=black] coordinates { (0,007)
 };
\addplot [
 boxplot prepared={
 lower whisker=0.000,lower quartile=1.000,
 median=2.000,
 upper quartile=4.000, upper whisker=8.000, box extend=0.6
}, mark=+, color=black!50] coordinates { (0,009)
(0,010)
 };
\addplot [
 boxplot prepared={
 lower whisker=12.000,lower quartile=21.000,
 median=25.000,
 upper quartile=27.000, upper whisker=35.000, box extend=0.6
}, mark=+, color=black] coordinates { (0,011)
 };
\addplot [
 boxplot prepared={
 lower whisker=0.000,lower quartile=1.000,
 median=4.000,
 upper quartile=7.000, upper whisker=16.000, box extend=0.6
}, mark=+, color=black!50] coordinates { (0,017)
(0,018)
(0,019)
 };
\addplot [
 boxplot prepared={
 lower whisker=1.000,lower quartile=8.000,
 median=11.000,
 upper quartile=13.000, upper whisker=20.000, box extend=0.6
}, mark=+, color=black] coordinates { (0,000)
(0,021)
(0,022)
(0,023)
(0,024)
(0,025)
 };
\addplot [
 boxplot prepared={
 lower whisker=0.000,lower quartile=8.000,
 median=11.000,
 upper quartile=15.000, upper whisker=25.000, box extend=0.6
}, mark=+, color=black!50] coordinates { (0,026)
(0,027)
(0,028)
(0,029)
(0,030)
(0,031)
(0,032)
(0,033)
(0,034)
(0,035)
(0,036)
(0,037)
 };
\addplot [
 boxplot prepared={
 lower whisker=22.000,lower quartile=42.000,
 median=51.000,
 upper quartile=56.000, upper whisker=76.000, box extend=0.6
}, mark=+, color=black] coordinates { (0,019)
(0,020)
 };
\addplot [
 boxplot prepared={
 lower whisker=14.000,lower quartile=50.000,
 median=64.000,
 upper quartile=74.000, upper whisker=110.000, box extend=0.6
}, mark=+, color=black!50] coordinates { (0,011)
(0,012)
(0,013)
 };
\end{axis}
\end{tikzpicture}

%% file: inc/NuScenesTrainval.tex
\pgfmathsetlengthmacro\MajorTickLength{
      \pgfkeysvalueof{/pgfplots/major tick length} * 0.5
    }
\pgfplotsset{
    every axis plot/.append style={fill,fill opacity=0.9},
    every boxplot/.style={mark=x,every mark/.append style={mark size=1.0pt}},
    every boxplot/.append style={
        boxplot/draw/lower whisker/.code={
            \draw[/pgfplots/boxplot/every lower whisker/.try] (boxplot cs:\pgfplotsboxplotvalue{lower whisker})
              -- (boxplot cs:\pgfplotsboxplotvalue{lower quartile});
        },
        boxplot/draw/upper whisker/.code={
            \draw[/pgfplots/boxplot/every upper whisker/.try] (boxplot cs:\pgfplotsboxplotvalue{upper quartile})
              -- (boxplot cs:\pgfplotsboxplotvalue{upper whisker});
              }
        },       
    boxplot/.cd,
        every lower whisker/.style={thick},
        every upper whisker/.style={thick},
        every median/.style={red,ultra thick}
}
\begin{tikzpicture}
\begin{axis}[
  y=0.25cm,
  xmin=-1,xmax=22,
  ytick=\empty,
  y dir=reverse,
  tick align=center, 
  major tick length=\MajorTickLength,
  yticklabel=\empty, 
  extra y ticks={1,1.5,2},
  extra y tick labels={,ANEES,},
  extra y tick style={major tick length=0pt},
  width=0.99\linewidth,
  name=ANEES, 
  legend columns=-1, 
  legend style={font=\footnotesize,draw=none,fill=none}, 
  every axis legend/.append style={at={(1.00,0.94)},anchor=south east}, 
  legend entries={MSM,MSM-D}
]
\addlegendimage{mark=square*,color=black,only marks, mark size=2pt}
\addlegendimage{mark=square*,color=blue,only marks, mark size=2pt}
\addplot [
 boxplot prepared={
 lower whisker=0.286,lower quartile=1.251,
 median=1.560,
 upper quartile=1.903, upper whisker=2.861, box extend=0.6
}, mark=+, color=black] coordinates { (0,0.196)
(0,0.263)
(0,2.939)
(0,2.987)
(0,2.994)
(0,3.037)
(0,3.136)
(0,3.232)
(0,3.251)
(0,3.330)
(0,3.371)
(0,3.451)
(0,3.465)
(0,3.596)
(0,3.631)
(0,3.665)
(0,3.673)
(0,3.804)
(0,3.898)
(0,3.940)
(0,4.366)
(0,4.379)
(0,6.022)
(0,9.979)
 };
\addplot [
 boxplot prepared={
 lower whisker=0.100,lower quartile=2.657,
 median=4.009,
 upper quartile=5.612, upper whisker=9.969, box extend=0.6
}, mark=+, color=blue] coordinates { (0,10.095)
(0,10.390)
(0,10.472)
(0,10.483)
(0,10.511)
(0,10.619)
(0,10.697)
(0,10.910)
(0,11.054)
(0,11.082)
(0,11.158)
(0,11.318)
(0,11.648)
(0,11.903)
(0,12.357)
(0,12.713)
(0,12.893)
(0,13.800)
(0,14.130)
(0,14.917)
(0,15.277)
(0,16.464)
(0,18.957)
 };
\end{axis}

\begin{axis}[
  y=0.25cm,
  xmin=-0.025,xmax=0.55,
  ytick=\empty,
  y dir=reverse,
  tick align=center, 
  major tick length=\MajorTickLength,
  yticklabel=\empty, 
  extra y ticks={1,1.5,2},
  extra y tick labels={,RMSE [m],},
  extra y tick style={major tick length=0pt},
  width=0.99\linewidth,
  name=RMSE,at=(ANEES.below south west), anchor=north west,
]
\addplot [
 boxplot prepared={
 lower whisker=0.090,lower quartile=0.141,
 median=0.159,
 upper quartile=0.178, upper whisker=0.233, box extend=0.6
}, mark=+, color=black] coordinates { (0,0.066)
(0,0.078)
(0,0.078)
(0,0.080)
(0,0.082)
(0,0.085)
(0,0.085)
(0,0.235)
(0,0.236)
(0,0.240)
(0,0.240)
(0,0.241)
(0,0.241)
(0,0.241)
(0,0.245)
(0,0.245)
(0,0.246)
(0,0.246)
(0,0.249)
(0,0.251)
(0,0.260)
(0,0.261)
(0,0.270)
(0,0.286)
(0,0.288)
(0,0.295)
(0,0.299)
(0,0.302)
(0,0.327)
(0,0.520)
 };
\addplot [
 boxplot prepared={
 lower whisker=0.005,lower quartile=0.010,
 median=0.012,
 upper quartile=0.013, upper whisker=0.019, box extend=0.6
}, mark=+, color=blue] coordinates { (0,0.000)
(0,0.001)
(0,0.001)
(0,0.001)
(0,0.001)
(0,0.001)
(0,0.001)
(0,0.002)
(0,0.002)
(0,0.002)
(0,0.002)
(0,0.002)
(0,0.002)
(0,0.002)
(0,0.002)
(0,0.002)
(0,0.002)
(0,0.002)
(0,0.002)
(0,0.003)
(0,0.003)
(0,0.019)
(0,0.019)
(0,0.019)
(0,0.019)
(0,0.019)
(0,0.019)
(0,0.020)
(0,0.020)
(0,0.020)
(0,0.021)
(0,0.021)
(0,0.021)
(0,0.021)
(0,0.022)
(0,0.027)
 };
\end{axis}
\begin{axis}[
  y=0.25cm,
  xmin=-0.025,xmax=0.55,
  ytick=\empty,
  y dir=reverse,
  tick align=center, 
  major tick length=\MajorTickLength,
  yticklabel=\empty, 
  extra y ticks={1,1.5,2},
  extra y tick labels={,RMSE [$^{\circ}$],},
  extra y tick style={major tick length=0pt},
  width=0.99\linewidth,
  name=RMSEA,at=(RMSE.below south west), anchor=north west,
]
\addplot [
 boxplot prepared={
 lower whisker=0.050,lower quartile=0.118,
 median=0.141,
 upper quartile=0.163, upper whisker=0.231, box extend=0.6
}, mark=+, color=black] coordinates { (0,0.026)
(0,0.029)
(0,0.045)
(0,0.235)
(0,0.238)
(0,0.241)
(0,0.250)
(0,0.255)
(0,0.258)
(0,0.261)
(0,0.262)
(0,0.268)
(0,0.270)
(0,0.288)
(0,0.299)
(0,0.308)
(0,0.320)
(0,0.322)
(0,0.330)
(0,0.354)
(0,0.453)
(0,0.472)
 };
 \addplot [
 boxplot prepared={
 lower whisker=0.044,lower quartile=0.080,
 median=0.093,
 upper quartile=0.105, upper whisker=0.141, box extend=0.6
}, mark=+, color=blue] coordinates { (0,0.019)
(0,0.028)
(0,0.034)
(0,0.035)
(0,0.036)
(0,0.036)
(0,0.039)
(0,0.039)
(0,0.040)
(0,0.042)
(0,0.144)
(0,0.147)
(0,0.148)
(0,0.149)
(0,0.151)
(0,0.152)
(0,0.154)
(0,0.160)
(0,0.162)
(0,0.164)
(0,0.185)
(0,0.199)
 };
\end{axis}
\end{tikzpicture}

%% file: inc/conclusion.tex

\section{Conclusion}

We presented a feasible way to solve the ego-motion estimation based on a thorough probabilistic strategy.
Utilizing the challenging automotive Doppler radar measurements with sparse environment representation and a high outlier rate, we were able to achieve robust estimation results, including a reasonable covariance quality.
Whereas the estimator's covariances in simulation showed a nearly perfect credibility of our algorithm -- released as open source\footnote{Available under \url{www.mytuc.org/creme}}.
Our extensive evaluation of about \SI{125}{\k} simulation experiments and about \SI{130}{\k} experiments based on the nuScenes dataset underlined the robustness of our method with and without using Doppler information.
Overall, it is perfectly suitable as simplified sensor information to be included within a multi-sensor fusion algorithm in a loosely-coupled way, as it is uncertainty aware.

As our method is based on standard least-squares minimization, but with more complex noise models representing the measurements and outliers, we think it is universal to be also used in related tasks.
These could be pure PSR on unstable point sets with known measurement noise or feature-based registration in 3D laser scans.
However, a remaining problem of optimization algorithms are local minima, which also lead to wrong covariance estimates. 
Even if our approach seems to be robust against this problem, further investigation can be done regarding this, e.g., by adding this possibility to the estimated covariance like done in \cite{Brossard20} by using the unscented transform.